# Communication-aware planning for robot teams deployment

Yaroslav Marchukov and Luis Montano

*Instituto de Investigación en Ingeniería de Aragón, University of Zaragoza, 50018 Zaragoza, Spain, (email: yamar,montano@unizar.es).*

**Abstract:** In the present work we address the problem of deploying a team of robots in a scenario where some locations of interest must be reached. Thus, a planning for a deployment is required, before sending the robots. The obstacles, the limited communication range, and the need of communicating to a base station, constrain the connectivity of the team and the deployment planning. We propose a method consisting of three algorithms: a distributed path planner to obtain communication-aware trajectories; a deployment planner providing dual-use of the robots, visiting primary goals and performing connectivity tasks; and a clustering algorithm to allocate the tasks to robots, and obtain the best goal visit order for the mission.

## 1. INTRODUCTION

The deployment of robot teams for exploration or environmental monitoring can be executed in many ways. When the goals are very dispersed and the connectivity between all the robots is not required, obviously it is better to distribute the team in order to cover the maximum area as soon as possible. In this case, the limited range of the signal and the obstacles prevent direct communication between robots, and this must be considered in the deployment planning. Works as (Hollinger and Singh (2012)) solve this problem planning paths to join the robots periodically sharing information during the mission. Others predict possible communication with independent agents and exploit this area to reach the goals (Flushing et al. (2014)). However, in both solutions, the robots spend an extra time for connectivity maintenance, rather than accomplishing the mission. More interesting way to control the connectivity among the robots is using them as relays, that is, providing communication to other members of the team (Stump et al. (2011), Fink et al. (2013), Tardioli et al. (2016)). The advantage of this type of deployment is the possibility to continuously communicate the team with a static control station, so that a human operator is able to monitor the execution of the mission. In this case, the team usually adopts a leader-followers approach, where one robot executes the mission and the rest of the team has the purpose of relay. The main drawback of this approach is produced in scenarios with very dispersed goals and in presence of many obstacles. It takes too much time to visit all the goals and the robots are wasted only for relay tasks.

When the robots are deploying, the obstacles are critical for connectivity maintenance. In (Stump et al. (2011)), the *line-of-sight* (LoS) is considered for the connections between the robots. This is the easiest way to assure the communication, but some valid positions are discarded even if the received signal is high enough for communication. Consequently, it increases the number of relays employed for the mission. Therefore, it is worth considering a *non-line-of-sight* (NLoS) component. Works (Rizzo et al. (2013), Yan and Mostofi (2012), Twigg et al. (2013)) accurately characterize the signal in the environment, considering the variations suffered by the signal in the propagation media.

The quality of communication is affected due to the propagation of the signal. This topic acquired great interest from research community. In works (Rizzo et al. (2013),Twigg et al. (2013)), the RSS (Received Signal Strength) was studied in order to assure good communication. Others, as (Fink et al. (2013), Yan and Mostofi (2012)), go further and evaluate the binary rate and bit error ratio, respectively. These methods produce reactive trajectories travelled by the robots, because of the complexity to predict these parameters.

Let us define the context of this work. Imagine for instance a scenario of an accident or CBRNE (Chemical, Biological, Radiological, Nuclear and Explosive) emergency in a building, where it is necessary to quickly localize and identify the critical places. A team of robots can be used to solve this situation for the first responders. In the present paper we develop a planner for a robot team deployment for reaching some places of interest. We assume that initially there are as many robots as places to be reached. But it is also possible to find a solution even when not all the robots are used. If the map of the scenario is available, a previous planning for a deployment before sending the robots can be carried out, although reactivity to signal coverage can be maintained. A human operator is monitoring the mission from a static control center, so when some robot achieves a goal, it must share information with the base station (BS). The mission has to be monitored as long as possible. However, a momentary disconnection during the motion is allowable, although the robots must be strictly connected in the goal positions. The direct communication with BS is impossible, thus the team is coordinated in order to enhance the coverage area. In contrast to leader-followers approach, we consider primary and relay tasks for every robot. The priority of the robots is to reach the goals first. Just after that, the robots can change their positions in order to serve as relays. At the same time, the robots should find direct paths to the goals when possible, to minimize the time of mission. Thus, the planner provides trajectories that can avoid or limit reactive motions.

The easiest way to deploy the team, it is planning the shortest path to every goal location. We employ the Fast

★ This research has been funded by projects DPI2012-32100 and DPI2016-76676-R-AEI/FEDER-UE of Spanish Government.

Marching Method (FMM), (Sethian (1995)) as the base method for path planning, as explained in Sect.4.1, because it fits well to be adapted for communication constraints planning. But the connectivity is not considered in that basic technique. Thus, the first contribution of this paper is a distributed communication-aware path planner, named as Communication-aware FMM (CA-FMM), which is employed by each robot to obtain the shortest path within coverage area, described in Sect.4.2. Using CA-FMM, the robots are deployed as independent agents which use the signal of other robots to communicate with BS. So, as a second contribution, we present a centralized deployment planner to coordinate the team deployment. A preliminary version of this algorithm was presented in (Marchukov and Montano (2016)), which obtains the possible connections for the team, the sequential coverage enhancement, and the optimal positions for the relays. Here, we extend this work, improving the relay task assignment and taking into account the difficulty to reach new relay positions in presence of obstacles. Furthermore, we employ a more accurate signal propagation model to obtain the coverage area, described in Sect.3. The joint use of CA-FMM with the deployment planner, called Deployment Planning FMM (DP-FMM), is described in Sect.5.1. If there are as number of available robots as goals (tasks) to be reached, the method ensures that the mission is executed in minimum time under conditions specified in Sect.5.2. But the objective might be to use the minimum number of robots as relays. For this case, the last contribution of this work is a centralized clustering algorithm, which allocates several goals to each robot and computes its optimal visit order. The use of this procedure with DP-FMM is named Deployment Planning and Allocation FMM (DPA-FMM), presented in Sect.5.2.

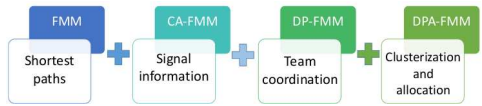

Fig. 1. Integration sequence of the algorithms

Each of the presented algorithms is an improvement of the previous one, Fig.1, and a simple example of the complete procedure is depicted in Fig.2. After obtaining the initial connectivity, Fig.2(a), only $g_{1-4}$ are connected with the BS, and can be reached by placing robots as relays at these positions. FMM computes the shortest paths without a complete connectivity, Fig.2(b). In Fig.2(c), CA-FMM obtains larger paths to $g_{1-4}$ deviating the robots towards coverage areas, but the goals $g_{4,5}$ are not reachable with communication. In Fig.2(d), the deployment planner computes new goal positions for relay tasks $g_{1',2'}$, to improve connectivity. These positions are the minimal to cover all the goals, as well as those that involve minimal displacement for the robots. Thus, after visiting $g_1$, the robot is considered free and it moves to serve as relay in $g_{1'}$. From this position it provides connectivity to $g_{2,3,4}$, thus the number of relays is reduced, and the shortest paths are within coverage area. Likewise, after reaching $g_2$, the robot moves to $g_{2'}$ to provide connectivity to reach $g_{5,6}$. DP-FMM employs all the robots in order to minimize the mission time, Fig.2(e). In Fig.2(f), DPA-FMM classifies the goals into 3 clusters $\{g_{4,3,2,2'}\},\{g_{1,1'}\},\{g_{5,6}\}$, so uses only 3 robots to accomplish the mission, at the expense of waste more time. The presented algorithms are also able to reactively respond to changes in the signal strength or obstacles in the environment. Metrics for the possible deployments, obtained with the presented algorithms, are presented in Sect.6.

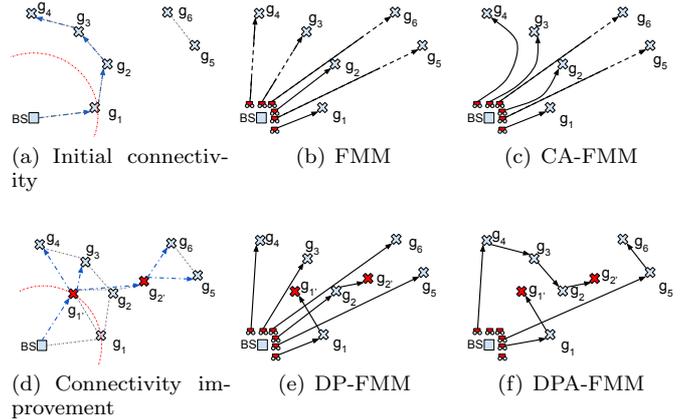

(a) Initial connectivity
(b) FMM
(c) CA-FMM
(d) Connectivity improvement
(e) DP-FMM
(f) DPA-FMM

Fig. 2. Deployment to visit 6 goals. In (a), the coverage area of BS is depicted with red circle, black dashed lines the possible connections, and blue arrows the chosen ones. The paths obtained by FMM and CA-FMM are represented in (b) and (c); the dashed lines represent the stretch of the path travelled without connectivity with the rest of the team. In (d), the deployment planner computes new positions to improve the connectivity, red crosses. The paths obtained by DP-FMM and DPA-FMM are depicted in (e) and (f), respectively.

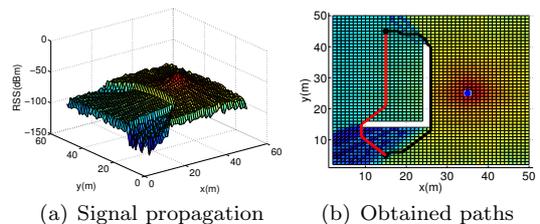

(a) Signal propagation
(b) Obtained paths

Fig. 3. Example of communication-aware paths. The signal propagated from a BS in $(35, 25)$ is depicted in (a). In (b), $x_r$ and $x_g$ are depicted with black circles, the position of BS is represented with blue circle. Considering only the distance, FMM obtains the shortest path, depicted as a red line. CA-FMM computes a larger path, depicted as a black line, that maximizes the stretch within the coverage area $\mathbf{x}_c$.

## 2. PROBLEM STATEMENT

### 2.1 Communication-aware paths

Consider an autonomous robot, with $x_r$ denoting its position. The mission of the robot is to reach a goal, located at $x_g$. There are some signal sources or transmitters present in the environment, whose coverage area can be defined as $\mathbf{x}_c : \{\mathbf{x} \mid RSS(\mathbf{x}) \geq \gamma\}$, where $\gamma$ is a $RSS$ threshold to assure communication. Let us define $\tau$ as a possible path from the initial position of the robot $x_r$ to the goal $x_r$. The standard FMM formulation provides the shortest path based in a wavefront propagation in all the directions using a propagation velocity $F$. The proposed technique modifies the velocity function $F$ in order to deviate the wavefront to areas within signal coverage areas. The result applying this new method, called CA-FMM, is depicted in Fig.3 and explained in detail in Sect.4. The path solution is not the shortest, but the one that avoid the obstacles and moves the robot through the communicated area.

### 2.2 Communication-aware deployment

Consider a team of N robots with $\mathbf{x}_r$ denoting their positions, $\mathbf{x}_{r_i} := (x_{r_0}, ..., x_{r_N})$. The team includes a static base station (BS), indexed with $i = 0$, which is able to communicate with the robots. The mission of the team is to visit N goals with $\mathbf{x}_g$ representing their locations, $\mathbf{x}_{g_i} := (x_{g_0}, ..., x_{g_N})$. Therefore, the team is self-coordinated by allocating every robot $i$ to every goal $i$. So the path between each pair of $x_{g_i}$ and $x_{r_i}$ is expressed as $\tau_i$.

Instead of a prior role assignment for the robots to primary and relay tasks, the team automatically distributes these roles. So all the robots can be used for both types of tasks. While the robots are moving they expand the coverage area within the environment. So it is necessary to compute the goal visitation order to know which robots are going to provide communication to others. We can define a graph, expressed as $\mathcal{G}(\mathbf{x}_g)$, whose vertexes are the goal positions and the edges are the connections between the goals. We consider a connection between positions $x_i$ and $x_j$ when $RSS(x_i, x_j) \geq \gamma$ and $RSS(x_j, x_i) \geq \gamma$. All those goals which are not connected to the BS, directly or through others, are not reachable with communication. The team should dedicate the minimal number of robots for relay tasks, so that greater number of robots are visiting primary goals. Therefore, given a graph $\mathcal{G}(\mathbf{x}_g)$, we compute a tree $T(\mathbf{x}_g)$, which minimizes the number of hops, or depth in the tree, from the root (BS) to every node. The number of hops to some position $x_{g_i}$ indicates the number of robots needed to reach this position with connectivity, and is expressed as $n(x_{g_i})$.

Only after visiting the primary goal, a robot is considered free and can change the position to enhance $\mathbf{x_c}$. Hence each robot attempt to find a new position where the maximum number of goals will be in its coverage area. In other words, the robot will provide communication to goals which were connected to another relay, thus the number of hops to these goals is reduced. At the same time, this procedure is applied to disconnected goals, in order to complete the mission with connectivity to all the goals when the robots reach them.

An example of this method is shown in Fig.2(a),2(d). The connections of $\mathcal{G}(\mathbf{x}_g)$ are depicted with black dashed lines, and $T(\mathbf{x}_g)$ is represented with blue arrows. In (d), after visiting $g_1$, a robot will move to $g_{1'}$, increasing the coverage area. In consequence, the number of required relays to reach $g_{3,4}$ will be reduced with respect to the situation in (a). The new position $g_{2'}$ will be used to connect the previously disconnected goals, $g_{5,6}$. As a result, the minimum number and locations of relays are computed in this step.

*2.3 Clustering*

The number of the available robots for the mission initially corresponds to the number of goals, so the fastest way to complete the mission is using all the robots. However, it is not always necessary, or possible, to employ the entire team. Thus, we study the minimization of the number of employed robots for achieving the proposed mission, which will be used to visit several goals. Firstly, the deployment planner has computed the relay locations, and allocates them to the robots. Then, the robots used for this purpose should remain in these positions, and the rest of the team can be freely used to reach other goals.

The first step is to find all those goals which can be visited by the same robot. This process of clustering is made depending on three factors: the initial point, the deviation distance from the straight path between goals, and the occupation at these positions. This can be observed in Fig.2(d). Possible destinations for the robots are $g_{1'}$ and $g_{2'}$, where they are occupied as relays. The rest of the goals only must be visited, in other words, it can be interpreted as waypoints to the destinations. Thus, the algorithm divides the goals into three different clusters, which will be visited by three robots. Once the clusters are obtained, each robot has multiple possibilities to visit the goals. Consequently, the goals visit sequence is computed in order to minimize the travelled distance. Fig.2(f) depicts the order that the robots follow for each cluster, visiting the goals. When the plan is already obtained, the path travelled by every robot is known by the rest of the teammates. So that, when a robot is aware that its relay will be delayed, it waits until the communication area is extended to its goal. For instance, in Fig 2(f), the robot which visits $g_{5-6}$ waits until the robot, which visits $g_{4-3-2-2'}$, reaches the last position, where provides coverage to $g_{5-6}$. This way, the travelled distance is minimized and at the same time of visiting the goals, the robots are connected with BS. The detailed procedure of clustering is explained in Sect. 5.2.

## 3. SIGNAL PROPAGATION

Every signal propagation depends on three main factors: the attenuation due to the distance, the shadowing because of obstacle traversing, and multipath, due to the reflexions of the signal. Thus, the RSS is obtained subtracting the path loss to the emitted power by the antenna. The path loss suffered by the signal can be defined as (Lott and Forkel (2001)):

$$L = L_o + 10n\log_{10}(d) + S + M \quad (1)$$

where $L_o$ is the path loss at distance of 1m, $n$ is the path-loss exponent, $d$ is the distance between transmitter and receiver, $S$ represents the fading due to the shadowing and $M$ the fading due to the multipath. The shadowing can be obtained as $S = n_w a_w$, where $n_w$ is the number of traversed walls and $a_w$ is the attenuation per traversed wall.

The attenuation and shadowing are easily predictable, but the multipath effect is computationally intractable using multiple mobile transmitters. Some advanced techniques, as ray-tracing, can be employed to obtain accurate approximations of the signal, but it is out of the scope of present work. Thus, we propose an approximation of the signal, which considers the main features of the signal propagation.

Our algorithm is aimed at using in indoor scenarios, emulating some commercial technology as WiFi. We have validated the model from the real signal data collected in a building at the University of Zaragoza. We consider LoS and NLoS components of the signal in order to increase the precision of our simulations. The building map and the collected signal data are depicted in Fig.4(a).

Firstly, we use the moving average to smooth the signal, depicted with blue line in Fig.4(b). Secondly, we subtract the average signal to raw data, obtaining the fast fadings of the signal produced by multipath effect. Extracting the variance, it is possible to approximate the multipath component by a Gaussian distribution $\mathcal{N}(0, \sigma_{mp}^2)$. As we have the RSS and the distance, the path-loss exponent $n$ is computed employing polynomial regression. Consequently, the polynomial coefficients are obtained with least-squares fit. The signal simulated using eq.(1), is presented in Fig.4(b), capturing the main shape of the real signal. This signal model has been used to obtain the connectivity between the teammates, by the deployment planner, as well as by the path planner developed in the following section.

## 4. COMMUNICATION-AWARE PATH PLANNING

In this section we develop the method to find paths leading to the goals with maximum connectivity, outlined

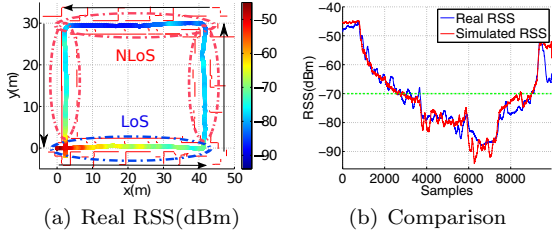

(a) Real RSS(dBm)  (b) Comparison

Fig. 4. RSS in a real environment. In (a), it is represented the collected data, the transmitter is located at (0,0), black arrows depict the walk through of the robot, LoS and NLoS areas surrounded with blue and red dashed lines, respectively. The comparison between smoothed real and simulated RSS, with eq.(1), is represented in (b). From (Rizzo et al. (2013)) we extract $\gamma = -70$dBm, depicted with green line. We take the values for shadowing of 10dB/wall and 2.5dB/glass plate, from (Lott and Forkel (2001)). The computed path-loss exponents are: $n_{LoS} = 1.7, n_{NLoS} = 1.4$; and multipath variances: $\sigma^2_{LoS} = 3.45, \sigma^2_{NLoS} = 3.25$.

in Sect.2.1. We briefly describe the FMM and present our CA-FMM approach.

### 4.1 Fast Marching Method (FMM)

The Fast Marching Method (FMM) was proposed as an approximated solution to Eikonal equation in Sethian (1995). It consists in the computation of a distance function $D$, from some source point for every point of the grid $\mathbf{x}$, obtaining as solution the minimum cost to reach these points. Thus, a wavefront is propagated over all the grid, computing the distance function $D$, with the metric $F$ representing the speed of the wavefront propagation. Notice that $F$ contains the obstacle information, so that $F(\mathbf{x}) = 0$ for all the positions of $\mathbf{x}$ which contain obstacles and $F(\mathbf{x}) = 1$ for free space. Then, the wave is initialized at goal $x_{g_i}$, assigning $D(x_{g_i}) = 0$, and it propagates uniformly in all the directions, computing:

$$|\nabla D|F(\mathbf{x}) = 1 \qquad (2)$$

The FMM can be interpreted as a continuous Dijkstra method. The main advantage is that FMM uses the values of two neighbours to interpolate the distance, instead of one, used by Dijkstra. Therefore, FMM obtains more accurate approximations of distance. As Dijkstra, FMM computes the shortest paths to the goals and not suboptimal solutions, as provided by randomized algorithms, such as different versions of RRTs (Rapidly exploring Random Trees).

### 4.2 Communication-aware FMM (CA-FMM)

Since the velocity function $F$ contains only obstacles information, the gradient $\nabla D$, is the Euclidean distance to the goal. However, our approach must include the signal information. So, the signal propagated by each relay, is combined with $F$, resulting in a new velocity function with communication $F_c$, defined as:

$$F_c(\mathbf{x}) = F(\mathbf{x}) + f(\mathbf{x}_c) \qquad (3)$$

where the function $f(\mathbf{x}_c)$ normalizes the signal of the coverage area and fix to zero the positions of the other robots, considered as obstacles. The new velocity function $F_c$, is used in eq.(2) to obtain gradients with a new metric, depending on the distance and communications. The wavefront is propagated faster in areas with higher values of $F_c$, which correspond to the coverage area. Therefore, lower values of the gradient of distance function $\nabla D$ computed using the new metric $F_c$, find shorter paths within coverage area.

**Algorithm 1:** CA-FMM (for one robot reaching one goal)

**Data:** Robot position $x_{r_i}$, Goal position $x_{g_i}$, Paths $\tau_l$ (travelled by the relays)
**Result:** Path $\tau_i$
1 **Function** $CA-FMM(x_{r_i}, x_{g_i}, \tau_l)$
2     **foreach** $\mathbf{x}_{r_l} \in \tau_l$ **do**
3        $\mathbf{x}_c = pathloss(\mathbf{x}, \mathbf{x}_{r_l}, \gamma, n)$ // Coverage area
4        $F_c \leftarrow compute(\mathbf{x}_c)$ // Using eq.(3)
5        $\nabla D \leftarrow propagate\_front(x_{g_i}, \mathbf{x}_{r_l}, F_c)$ // Using eq.(2)
6     **end**
7     $\tau_i \leftarrow gradient\_descent(\nabla D)$
8 **end**

As the signal sources are mobile, the CA-FMM (Alg.1) takes into account the movement of the robots and the variations of the signal. Firstly, the algorithm obtains the coverage area $\mathbf{x}_c$ (l.3) and this information is inserted into $F_c$, (l.4). The coverage area of is computed employing eq.(1), but it can be obtained using the real signal data as well. Then, the wavefront is propagated to the relay positions with every movement of the relays (l.5). The algorithm iterates until reach the goal position, using the coverage of already deployed robots and finally it constructs the path (l.7).

## 5. DEPLOYMENT PLANNING

In the present section we present our method to assign coverage tasks to each robot devoted to do it, and to compute the clusters and the sequence of visit of the primary goals in each of them.

### 5.1 Sequential deployment and coverage enhancement (DP-FMM)

When some robot is going towards its goal, it modifies the coverage area, providing signal to other robots. As explained in Sect.2.2, the connectivity tree $T(\mathbf{x}_g)$ provides the connections between the goals and the depth of each one. Hence, the order of goal visitation is obtained to ensure that the coverage is sequentially enhanced. Besides the sequence, we are able to know which robot must remain at the same position providing connectivity. This process is illustrated in Fig.2(a). When a robot achieves the position of $g_1$ enables $g_2$ and so on. Until a robot reaches $g_4$ the relays must remain in $g_{1-3}$. Once the primary goals are visited, the deployment planner computes new goal locations where the robots, serving as relays, improve the connectivity of the team in terms of distance and required relays. The number of these new goal positions is the minimal, thus it involves the minimal number of robots devoted to relay tasks. Simultaneously, the tasks are allocated to those robots, which require minimal displacement to relay positions. From Fig.2(a), we extract $d(\tau(x_{g_4})) = d_{01} + d_{12} + d_{23} + d_{34}$ and the depth is $n(x_{g_4}) = 4$. In Fig.2(d), after reaching $g_1$, the robot moves to $g_{1'}$, obtaining $d(\tau(x_{g_4})) = d_{04}$, considerably smaller than initially, and $n(x_{g_4}) = 2$. At the same time, the second robot enhance the coverage area at $g_{2'}$, connecting $g_5$ and $g_6$.

The deployment planner (DP) assures the connectivity to all the goals if the condition $d(\tau_f)/d_{cov} \leq N$ is accomplished, where $\tau_f$ is the minimum distance path from the BS to the farthest goal (in the sense of the shortest path computed, for instance, by an $A^*$ algorithm), $d_{cov}$ is the minimum coverage distance where the connectivity with a robot is always assured, and $N$ is the number of robots. Note that in the worst case the robots form a chain up to the farthest goal, so under that condition all the goals will be always reached.

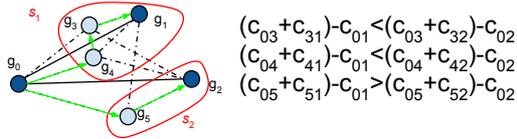

Fig. 5. Clustering and goal assignment procedure for 2 robots visiting 5 goals, $g_{1-5}$. The visit sequence for each cluster $s_1, s_2$ is depicted with green arrows.

This preliminary deployment algorithm was presented in Marchukov and Montano (2016). Here, we have improved the relay task allocation. In the previous work, when the new relay positions were computed, the *idle* robots were the first to visit them. That is, the robots which were not used for relay tasks in that moment. As a result, the nearby relay robots were automatically discarded, even being optimal for these tasks. Now, we use the distance as the cost to reach some goal. In addition, we incorporate a penalty distance which corresponds to the difficulty to surround the obstacles between the robot and the goal. Therefore, the cost to go to some position $j$ from position $i$ is:

$$c(x_i, x_j) = d(x_i, x_j)(1 + n_o(x_i, x_j)) \quad (4)$$

where $d$ is the Euclidean distance and $n_o$ represents the number of obstacles in between $x_i$ and $x_j$. More sophisticated cost functions could be used for taking into account obstacle characteristics (size, material, etc.), but it does not change the essence of the basic algorithm proposed.

The Hungarian algorithm is used to allocate the tasks according to the computed costs. The deployment planning algorithm working altogether with the previously described, CA-FMM, is named as DP-FMM. At first, the centralized deployment planner obtains the tree, the list of first goals to visit and the new positions for relay tasks. This information is shared with the robots. Each robot computes the path using the distributed CA-FMM, sharing this information with the team. Note that in the first iteration, the coverage area is provided only by BS. The robots use the coverage area of their mates in each iteration. The algorithm iterates until obtaining the paths to all the goals, Fig. 2(d)-2(e).

### 5.2 Clustering and visit sequence (DPA-FMM)

The previous algorithm can minimize the mission time, because all the robots of the team are used, planning the shortest paths within the coverage areas. By contrast, we could want to minimize the number of robots required to complete the mission, instead of minimizing the mission time, as described in Sect.2.3. As a result, the number of goals visited by each robot is maximized. To this end, it is necessary to compute which goals should visit each robot. Thus, the goals are clustered according to different parameters and each cluster will be visited by only one robot, minimizing the total number of used robots.

We are going to group the goals connected to the same relay, where some robot is providing signal to the rest of the goals. The robots will come from this direction, following the trace of the signal, so we consider it as the starting point, $g_0$ in Fig.5. The number of clusters is determined by the number of relay goals to achieve. That is, if there are two goals where the robots will have to remain providing connectivity, at least two robots are required, and these goals are the destinations of those robots, $g_{1,2}$ in the figure. All those goals which are not used for relay tasks are considered waypoints for one robot to a destination, $g_{3-5}$. The waypoints included in that cluster, will be those that will provide the smallest deviation distance with respect to the direct path to the destination point of this cluster. Mathematically, the clustering process may be expressed as,

$$s_i = \{x_p : (c_{lp} + c_{pi}) - c_{li} < (c_{lp} + c_{pj}) - c_{lj} \\ \forall j, 1 \geq j \geq k\}, \quad (5)$$

where $c$ represents the cost computed with eq.(4), $k$ is the number of destinations, $i$ and $j$ represent two possible destinations, $p$ is a waypoint and $l$ is the link of $p$ and $k$. An example of this procedure is shown in Fig.5.

After obtaining the clusters, it is necessary to assign the best order to visit the waypoints. Given the limited signal range of the relays, the dispersion of the goals and the presence of several destination points, the clustering process greatly reduces the computational cost in the process of obtaining the visit sequence. Hence we assess all the possible combinations of visit order of the waypoints from starting point to the destination in each cluster. The optimal sequence is that which provides the shortest distance, Fig.5. The distance cost between every pair of points is computed using eq.(4).

The DP-FMM including the goal allocation receives the name of DPA-FMM and the procedure is shown in Alg.2. At each iteration the deployment planner obtains the goals to visit and the new positions for relay tasks (l.2). The goals are assigned to each robot (l.5) and this information is shared with the robots. Each robot obtains the best visit order, computes its path using CA-FMM (l.6-10) and shares this information with the team. This solution minimizes the number of robots employed for the mission, as well as the total distance travelled by the team.

While the team is executing the mission, the robots can detect variations of the environment, as well as handle with variation of the signal. Therefore, our planner is used to replan in case of changes in the scenario, reactively solving this kind of situations. The only difference is that the robots which have reached their goals are extracted from the list of goals to visit (l.4), being the planner DPA-FMM launched again.

Every robot computes the shortest path under communication constraints, as defined in Sect.2.1. So that, the distance of the obtained path is considered the minimum. The farthest goal location constraints the time of the mission. Therefore, the condition to execute the mission in minimum time, is to use all the robots, thus each robot reaches its own goal. In the Fig.2, $g_6$ is the restrictive goal. As DP-FMM considers all the robots for the deployment, the distance is the direct distance to this goal, $d(\tau(x_{g6})) = d_{06}$ in Fig.2(e). Employing DPA-FMM the robots visit several goals. From Fig.2(f), we extract $d(\tau(x_{g6})) = d_{05} + d_{56}$. Clearly, the time is increased with respect to DP-FMM solution, at the expense of reduce the number of robots.

## 6. RESULTS

We have evaluated the method by means of simulations, in order to test its effectiveness [1]. Thus, we test the four described algorithms for the deployment: FMM, CA-FMM, DP-FMM and DPA-FMM. The results are evaluated according to different metrics: travelled distance ($d$), mission time ($T$), connectivity ($C$), and robot occupation during

---

[1] The video can be found in
http://robots.unizar.es/data/videos/ifac17yamar.mp4

**Algorithm 2:** DPA-FMM

    **Data:** Robot positions $\mathbf{x}_r$, Goal positions $\mathbf{x}_g$,
            Reached goals $\mathbf{f}$
    **Result:** Paths $\tau$
1  **Function** $DPA-FMM(\mathbf{x}_r, \mathbf{x}_g)$
2     $T(\mathbf{x}_g) \leftarrow DP(\mathbf{x}_g)$ // deployment planner
3     **for** $dp = 1,..,max(n(\mathbf{x}_g))$ **do**
4        $\mathbf{x}_g(dp) \leftarrow \mathbf{x}_g(dp) \setminus \mathbf{x}_g(\mathbf{f})$ // $\mathbf{x}_g(\mathbf{f})$ represents already planned goals
5        $\mathbf{S} \leftarrow cluster(\mathbf{x}_g(dp))$ // $\mathbf{S}$ represents the clusters of the same tree depth
6        **foreach** $s_i \in \mathbf{S}$ **do**
          // $s_i : \{x_{r_i}, x_{g_{w_1}}, ..., x_{g_{dest_i}}\}$, robot position, waypoints, destination
7           **foreach** $x_j \in s_i$ **do**
8              $\tau_i \leftarrow CA-FMM(x_j, x_{j+1}, \tau_{r_l})$
               // $r_l$ is the relay of the goals of $s_i$
9           **end**
10       **end**
11       $T(\mathbf{x}_g) \leftarrow DP(\mathbf{x}_g)$ // tree update for new relay positions
12    **end**
13 **end**

the mission ($O$). The last two metrics correspond to communication of the team, the connectivity represents the time which all the team remain connected to BS. The occupation represents the time employed by every robot to provide connectivity to the rest of the team. The results of time, connectivity and occupation are normalized to the maximum value of all the algorithms. $C_{mean}$ and $O_{mean}$ represent the average time in which a robot is connected and employed as relay, respectively. The worst cases of distance and connectivity are $d_{max}$ and $C_{min}$, respectively. $d_{tot}$ is the sum of travelled distance by the team.

An example of the deployments obtained by each algorithm is depicted in Fig.6, and numerical results are presented in Table 1. We use an Hungarian algorithm for task allocation in FMM and CA-FMM, the costs computed by eq.(4). FMM obtains the shortest paths, because it considers only the distance for planning, Fig.6(a). CA-FMM deviates the robots to areas where connectivity is maximized, Fig.6(b), although it is not able to finish the mission reaching all the goals with connectivity. In Fig.6(c), DP-FMM computes and allocates the new positions for relay tasks, thus the mission is accomplished with complete connectivity for the whole team. This involves a delay of 6%, and 14.7% of extra travelled distance (Table.1) with respect to the basic solution without connectivity issues (FMM). Employing the complete planner DPA-FMM, the mission is accomplished reducing 23% the total distance travelled by the team. As a result, the longest distance travelled by a robot, which delimits the time of the mission, is increased in 31% with respect to the FMM algorithm, but the number of used robots is strongly reduced, 5 instead of 10. Moreover, some robots are temporarily disconnected from the rest of the team during the motion. This occurs when a robot visits several goals and considerably deviates from the direct path to its relay. Consequently, the robots that were connected to this robot are disconnected too. This situation is shown in Fig.6(d), losing the connectivity only during 3% of time, Fig.6(g). The method allows disconnections during the motion between the goals, but the robots always establish communication when they visit the goals and BS shares information (see the video). Due to the deployment planner, both, DP-FMM and DPA-FMM perform better allocation of relay tasks. Therefore, the occupation is concentrated in robots which are the best for relay task.

Table 1. Obtained results for the scenario represented in Fig.6.

|  | $d_{max}$ | $d_{tot}$ | $T$ | $C_{mean}$ | $C_{min}$ | $O_{mean}$ |
|---|---|---|---|---|---|---|
| **FMM** | 49.36 | 310.26 | 0.47 | 0.79 | 0.70 | 0.24 |
| **CA-FMM** | 51.11 | 315.43 | 0.50 | 0.91 | 0.84 | 0.53 |
| **DP-FMM** | 53.70 | 363.75 | 0.53 | 1 | 1 | 0.33 |
| **DPA-FMM** | 71.50 | 225.58 | 1 | 0.99 | 0.97 | 0.50 |

Table 2. Obtained results for the scenario represented in Fig.7. Now DPA-FMM requires to use 6 robots instead of 5 of the initial plan.

|  | $d_{max}$ | $d_{tot}$ | $T$ | $C_{mean}$ | $C_{min}$ | $O_{mean}$ |
|---|---|---|---|---|---|---|
| **FMM** | 59.46 | 338.52 | 0.74 | 0.92 | 0.53 | 0.42 |
| **CA-FMM** | 59.18 | 345.63 | 0.72 | 0.84 | 0.42 | 0.49 |
| **DP-FMM** | 58.8 | 379.06 | 0.74 | 0.99 | 0.98 | 0.45 |
| **DPA-FMM** | 83.77 | 234.31 | 1 | 0.93 | 0.65 | 0.57 |

Table 3. Average results visiting different number of goals $\mathbf{N}$.

| N | Algorithm | $d_{max}$ | $d_{tot}$ | $T$ | $C_{mean}$ | $C_{min}$ | $O_{mean}$ | $R$ |
|---|---|---|---|---|---|---|---|---|
| 15 | **FMM** | 57.7411 | 515.2154 | 0.6503 | 0.7492 | 0.4511 | 0.2164 | 15 |
|  | **CA-FMM** | 62.5775 | 533.6519 | 0.7088 | 0.7365 | 0.4634 | 0.2211 | 15 |
|  | **DP-FMM** | 68.2503 | 599.3198 | 0.7682 | 0.9827 | 0.8501 | 0.2795 | 15 |
|  | **DPA-FMM** | 85.6530 | 488.6692 | 1.0000 | 0.9745 | 0.8051 | 0.4037 | 11 |
| 20 | **FMM** | 61.1598 | 714.9101 | 0.6927 | 0.8392 | 0.5535 | 0.2254 | 20 |
|  | **CA-FMM** | 66.8139 | 746.3166 | 0.7453 | 0.8417 | 0.5268 | 0.2376 | 20 |
|  | **DP-FMM** | 69.8790 | 819.1511 | 0.8029 | 0.9879 | 0.8818 | 0.2597 | 20 |
|  | **DPA-FMM** | 78.4160 | 583.9208 | 1.0000 | 0.9846 | 0.8700 | 0.4115 | 13 |
| 30 | **FMM** | 61.3190 | 1050.02 | 0.6715 | 0.9172 | 0.7536 | 0.1907 | 30 |
|  | **CA-FMM** | 66.4488 | 1093.3 | 0.7475 | 0.9230 | 0.7435 | 0.2063 | 30 |
|  | **DP-FMM** | 69.6423 | 1162.5 | 0.7845 | 0.9964 | 0.9372 | 0.2104 | 30 |
|  | **DPA-FMM** | 85.9442 | 752.3 | 1.0000 | 0.9880 | 0.8690 | 0.3803 | 17 |

The planner handles with variations when the team is deploying when the plan is executed. If the robot detects new obstacles, the signal changes and the planner re-plan new paths, resolving reactively these changing situations. The new paths are depicted in Fig.7 and the results are shown in Table.2. The distances and the time increase, and the connectivity, particularly $C_{min}$, is reduced. In case of FMM, $C_{mean}$ increases because the team subconsciously deviates to coverage areas. Now, DP-FMM loses 1% of connectivity in areas between goals, and DPA-FMM employs an extra robot than for the initial plan.

Table 3 depicts the average results in 50 simulations for each scenario, in missions visiting 15, 20 and 30 randomly distributed goals for different number of robots $R$. We can observe that the results follow the trend of the Table 1. Using the complete algorithm DPA-FMM, the mission is accomplished without employing the entire team, reducing the ratio robots-goals; and travelling shorter distance $d_{tot}$. But in exchange, the time of the mission is increased. The worst cases of connectivity for DP-FMM and DPA-FMM, are always above 80% of time, considered assumable. In absence of deployment planner, CA-FMM may obtain worse connectivity than for FMM, as for 20 robots case. This occurs when a robot is using the coverage area of an independent relay, which is accomplishing its own task. Thus, it is deviated from the goal to finally travel the remaining path without connectivity.

## 7. CONCLUSIONS AND FUTURE WORK

In this paper we present a method to plan a deployment of a team of mobile robots, to visit some places of interest within an environment with obstacles and under communication constraints, using some robots in role of relay. The team communicates with a base station, at the moment of visiting the goals. The plan is obtained previously to deploy the team, using only a map of the environment and a signal propagation model. But the method can reactively respond to environment or signal changes during the deployment. The presented method is integrated by three algorithms: a communication-aware path planner (CA-FMM), a deployment planner (DP-FMM), and a robot and goals clustering algorithm (DPA-FMM). With DP-FMM, the team fulfills the mission employing all the robots, so that reducing the time. In contrast, DPA-FMM allocates several goals per robot, so the amount of required members is reduced.

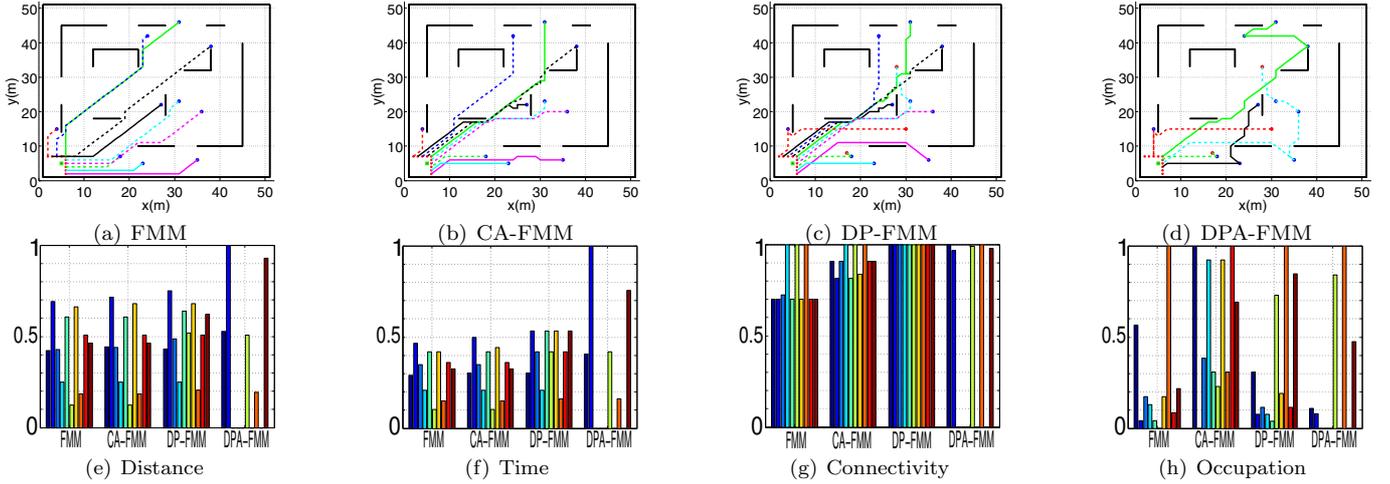

Fig. 6. Possible deployments visiting 10 randomly distributed goals. The paths obtained by each algorithm are illustrated in (a)-(d). Green square depicts the BS, red crosses represent the initial robot positions, blue and red circles are the primary and relay goals, respectively. The normalized results for the paths of the team are shown in (e)-(h) for the four algorithms. DPA-FMM employs 5 robots for the mission, thus only 5 bars are depicted.

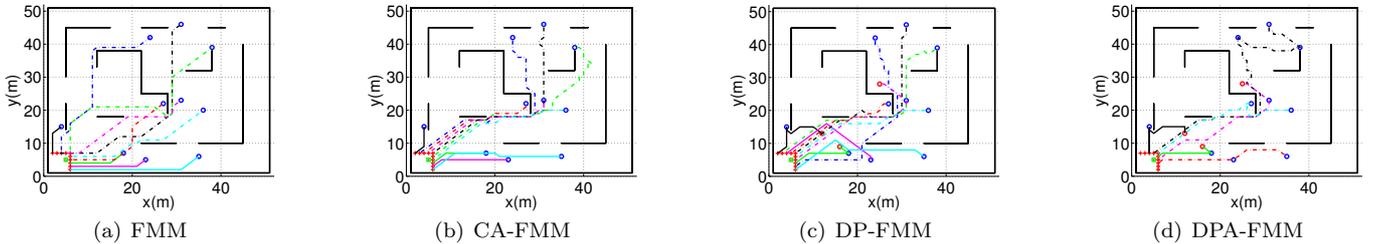

Fig. 7. New deployments after changes in the environment. During the deployment, a new obstacle is detected. The way is now blocked at the center of the scenario, and the signal also changes. The planner is re-launched from this new situation to achieve the rest of the tasks.

The problem of one robot visiting several objectives may deviate the robots from their relay tasks. Since some robots are used in role of relay, these deviations may temporarily disconnect some robots, as well as delay the mission. Currently, we are working in a distributed method to recover a member, which is suddenly disconnected from the team, due to a harsh signal fading as well as a drastic scenario variation. Also, we want to deepen in goals allocation considering the occupation of the robots for relay tasks during the motion. Thus, we propose to employ some modified version of MTSP (Multiple Travelling Salesman Problem). We want to adapt our deployment algorithm for a specified number of available robots, lower than number of goals. Moreover, we want to extend the algorithm to automatically fit the signal propagation parameters, while the robots are deploying. Real-world experimental assessment will be carried out.